\documentclass[a4paper,fleqn]{cas-dc}

\usepackage[authoryear,longnamesfirst]{natbib}

\def\tsc#1{\csdef{#1}{\textsc{\lowercase{#1}}\xspace}}
\tsc{WGM}
\tsc{QE}
\tsc{EP}
\tsc{PMS}
\tsc{BEC}
\tsc{DE}
\begin{document}
\let\WriteBookmarks\relax
\def\floatpagepagefraction{1}
\def\textpagefraction{.001}

\shorttitle{Revolutionizing Mobile Interaction: Enabling a 3 Billion Parameter GPT LLM on Mobile}

\shortauthors{Tomás Marques, Samuel Carreira}

\title [mode = title]{Revolutionizing Mobile Interaction: Enabling a 3 Billion Parameter GPT LLM on Mobile}                      

\tnotetext[1]{This document is the result of the practical project for Context Aware Systems of the masters in mobile computing.}

\author[1]{Tomás Marques}[type=editor,
                        linkedin=tomás-marques-19b531142,
                        orcid=0009-0002-9783-9752,
                        role=Author]

\ead{tomas.g.marques@ipleiria.pt}


\credit{Conceptualization of this study, Methodology, Software}

\affiliation[1]{organization={IPLeiria},
    addressline={ Morro do Lena, Alto do Vieiro}, 
    city={Leiria},
    % citysep={}, % Uncomment if no comma needed between city and postcode
    postcode={2411-901 PT}, 
    country={Portugal}}

\author[1]{Samuel Carreira}[%
    type=editor,
    role=Author,
    linkedin=samuel-vitorino-871113159,
    orcid=0009-0001-2117-7994
    ]

\ead{samuel.s.carreira@ipleiria.pt}

% Thrid Author

\author[1]{Carlos Grilo}[%
    type=editor,
    role=Author,
    orcid=0000-0001-9727-905X
    ]

\ead{carlos.grilo@ipleiria.pt}

\author[1]{José Ribeiro}[%
    type=editor,
    role=Author,
    orcid=0000-0003-3019-1330
    ]

\ead{jose.ribeiro@ipleiria.pt}

\begin{abstract}
The field of Artificial Intelligence has witnessed remarkable progress in recent years, especially with the emergence of powerful large language models (LLMs) based on the transformer architecture. Cloud-based LLMs, such as OpenAI's ChatGPT, offer impressive capabilities but come with concerns regarding latency and privacy due to network dependencies. This article presents an innovative approach to LLM inference, envisioning a future where LLMs with billions of parameters can be executed directly on mobile devices without network connectivity. The article showcases a fine-tuned GPT LLM with 3 billion parameters that can operate smoothly on devices with as low as 4GB of memory. Through the integration of native code and model quantization techniques, the application not only serves as a general-purpose assistant but also facilitates seamless mobile interactions with text-to-actions features. The article provides insights into the training pipeline, implementation details, test results, and future directions of on-device LLM inference. This breakthrough technology opens up possibilities for empowering users with sophisticated AI capabilities while preserving their privacy and eliminating latency concerns.
\end{abstract}

% \includegraphics{figs/grabs.pdf}

% Keywords

\maketitle

\section{Introduction}

Over the past few years, the field of Artificial Intelligence has experienced a remarkable evolution, with significant advancements in large language models (LLMs). The emergence of the transformer architecture in 2017 \citep{vaswani2017attention}, particularly in the case of Generative Pre-trained Transformers (GPT), has revolutionized the landscape. One notable example is ChatGPT, developed by OpenAI \citep{openai}, which has garnered widespread attention and found its way into numerous applications flooding major app stores.

Leveraging powerful cloud servers for intensive processing tasks offers numerous advantages, similar to the approach employed by OpenAI and other leading organizations. One notable benefit is the ability to achieve faster inference capabilities. Cloud servers with robust computational resources can process complex tasks more efficiently, allowing for quicker results and improved performance. However, recognizing the substantial disadvantages of this strategy is crucial. Potential latency problems are one noteworthy area of worry as results might not be available right away since data must be sent over the network to the cloud server for processing. In situations or applications where real-time responses are necessary or time-sensitive, this latency might be troublesome. Additionally, privacy concerns arise from the necessity of transmitting all information over the network to the cloud server. This means that sensitive data may be exposed to potential security risks during transmission. Protecting user privacy and ensuring the security of data are critical considerations, particularly in applications involving personal or confidential information. Given these factors, an ideal scenario would involve running the model offline directly on the device itself. This approach mitigates the concerns regarding latency and privacy, as the processing occurs locally without the need to transmit data over the network. By executing the model directly on the device, users can benefit from faster inference times while maintaining control over their data and preserving their privacy.

In this article, we present a groundbreaking approach to performing inference on GPT LLMs that we believe represents the future of this technology. With the increasing prevalence of custom accelerators in mobile devices, we envision a future where real-time inference of massive LLMs with billions of parameters can be executed directly on the device, without the need for a network connection. While we are still in the early stages, our application provides a promising glimpse into the immense potential of running LLMs strictly on the edge device.

Our application showcases the capabilities of a fine-tuned GPT LLM with 3 billion parameters, which can run smoothly on devices with as low as 4GB of memory. This remarkable achievement is made possible through our integration of native code and model quantization techniques. Not only that, but we go a step further: our refined model not only performs well as a general-purpose assistant but also facilitates seamless mobile phone interactions, enabling users to carry out various tasks including making calls, conducting web searches, and adding appointments to the calendar. Additionally, from the automatic model download to its utilization, we have built the application to be user-friendly, removing the requirement for prior AI knowledge.

This article's structure is as follows: Section~\ref{sec:2} provides a summary of the current state-of-the-art in LLMs, including methods, frameworks, and libraries. Section~\ref{sec:3} delves into our training pipeline, covering aspects such as the base model selection, dataset, fine-tuning, and quantization. Section~\ref{sec:4} provides detailed insights into the implementation of the mobile application, including how we interface with native code, the user interface screens, our innovative text-to-actions feature, and any limitations we encountered during development. In Section~\ref{sec:5}, we present the results of several tests conducted on the application, examining the model sizes before and after quantization, as well as showcasing sample prompts and their outputs from the final model.

Finally, in Section~\ref{sec:6}, we conclude with reflections on our work and discuss the future directions of this technology. The potential for on-device inference of LLMs is vast, and we believe it will revolutionize the way we interact with AI systems. With further advancements and optimizations, we envision a future where powerful LLMs are accessible and seamlessly integrated into everyday devices, empowering users with sophisticated AI capabilities directly at their fingertips.

\section{State of the art}
\label{sec:2}

Numerous projects in the area of large language models have emerged, offering frameworks, methods and platforms that greatly facilitate the development and deployment of chat applications. This section highlights a selection of these influential projects, that are making contributions to this area of study.

\subsection{OpenAI GPT model family}

The OpenAI GPT (Generative Pre-trained Transformer) model family represents a significant milestone in natural language processing and artificial intelligence research. Developed by OpenAI, the GPT models have garnered immense attention due to their ability to generate coherent and contextually relevant text.

The latest iterations of the GPT model family, GPT-3.5 and GPT4 \citep{openai2023gpt4}, serve as the foundation for the language capabilities of ChatGPT. Building upon the success of its predecessors, GPT-3.5, GPT-3, and GPT-2, GPT-4 incorporates several advancements to improve performance and address certain limitations.

One notable aspect of the GPT models is their transformer architecture. Based on the groundbreaking Transformer architecture, the GPT models leverage self-attention mechanisms to capture contextual dependencies and generate text that closely matches the given input. This architecture allows the models to effectively handle various natural language processing tasks.

Training the GPT models involves pre-training on a large corpus of text data, followed by fine-tuning on specific downstream tasks. OpenAI employs a massive dataset encompassing a wide range of sources, such as books, articles, and websites, to provide a broad understanding of language to the models during pre-training. Fine-tuning further refines the models' performance on specific tasks, making them more adaptable and applicable to various domains.

The performance of the OpenAI GPT models has been a subject of awe and admiration. GPT-3.5, in particular, exhibits remarkable language generation capabilities. It can produce coherent and contextually relevant responses to prompts, showcase a rudimentary understanding of complex topics, and mimic the writing style of specific authors or genres.

\subsection{Quantization}

Quantization \citep{lin2023awq} refers to the process of reducing the precision or bit-width of numerical values in a model or dataset. In the context of large language models like GPT-3, quantization can be applied to reduce the memory footprint and computational requirements of the model, making it more efficient to deploy and run on different hardware devices.

Large language models often use 32-bit floating-point values (float32) to represent weights, activations, and gradients. 
These high-precision representations require more memory and computational resources. However, by employing quantization, these values can be converted into lower precision formats such as 8-bit integers or 16-bit floating-point (float16), reducing the memory and processing power demands.
The process of quantization typically consists of two steps: calibration and quantization. During calibration, the model is run on a representative dataset to collect statistics about the range and distribution of values. These statistics are used to determine appropriate ranges and scales for quantization. The quantization step then maps the original high-precision values to their lower precision counterparts based on the determined scales and ranges.

A Large language model's memory footprint and processing needs can be reduced through quantization, allowing it to be used on devices with limited resources, such as mobile phones or embedded computers. However, quantization also introduces some precision loss, which may have a slight negative impact on the performance of the model. Therefore, a balance must be struck between the reduction in resource requirements and the impact on model accuracy when applying quantization techniques.Low-precision training is a technique used to train neural networks with reduced numerical precision for both weights and activations. By using lower precision, typically 16-bit or even lower, instead of the standard 32-bit floating-point representation, low-precision training aims to achieve more efficient and faster training while minimizing memory usage and computational resources.

\subsection{LoRA}

Traditional fine-tuning or adaptation of large language models involves updating all the parameters of the model, which can be computationally expensive and resource-intensive. LoRA \citep{hu2021lora} offers an alternative approach by learning pairs of rank-decomposition matrices while keeping the original weights frozen. This reduces the number of trainable parameters and, consequently, the storage requirements.

By using rank-decomposition matrices, LoRA represents the weights of the language model in a compressed format. These matrices capture the essential information needed for adaptation while discarding less relevant details. This compression significantly reduces the memory footprint required to store the adapted model.

The reduced storage requirement of LoRA has several benefits for efficient training. It allows for more effective utilization of computational resources, enabling larger models to be trained on limited hardware. Additionally, the reduced memory footprint facilitates faster data loading during training, leading to shorter training times.

LoRA also allows for low-precision training which is a technique used to train neural networks with reduced numerical precision for both weights and activations. By using lower precision, typically 16-bit or even lower, instead of the standard 32-bit floating-point representation, low-precision training aims to achieve more efficient and faster training while minimizing memory usage and computational resources.

\subsection{PEFT}

One framework that has garnered significant attention for efficient model fine-tuning is the PEFT (Parallel Environment Fine-Tuning) library. PEFT \citep{huggingface-peft} is designed to optimize and accelerate the training and fine-tuning of deep learning models in a distributed environment. It builds upon state-of-the-art techniques in parallel computing and offers several key features that contribute to its effectiveness and versatility.

\begin{itemize}
    \item Scalability and Performance: The PEFT library focuses on maximizing the scalability and performance of distributed deep learning. It provides efficient communication patterns and algorithms that minimize overheads and latency in data synchronization and parameter updates across multiple devices or nodes. By leveraging techniques like model parallelism and data parallelism, PEFT can efficiently distribute and synchronize the computation, resulting in faster convergence and improved training throughput.
    \item Flexibility and Compatibility: PEFT is designed to be compatible with various deep learning frameworks, such as TensorFlow and PyTorch. This compatibility enables researchers and practitioners to leverage existing models, codebases, and pre-trained models without significant modifications. By providing a unified interface, PEFT simplifies the process of distributed training and fine-tuning, allowing users to seamlessly integrate it into their existing workflows.
    \item Fault Tolerance and Resilience: Distributed deep learning systems often face challenges related to hardware failures, network issues, or node dropout. The PEFT library incorporates fault tolerance mechanisms that enhance the system's resilience to such failures. It includes techniques for checkpointing, model replication, and dynamic resource allocation to ensure that training processes can recover from failures and continue seamlessly, minimizing the impact on overall training performance.
    \item Performance Optimization: PEFT incorporates a range of performance optimization strategies to enhance the efficiency of distributed deep learning. It provides automatic graph partitioning algorithms that optimize the distribution of model layers across multiple devices, taking into account communication costs and device capabilities. Additionally, PEFT employs advanced gradient aggregation techniques and efficient data shuffling algorithms to further reduce communication overhead and improve training speed.
    \item Ease of Deployment: The PEFT library offers a user-friendly interface and straightforward deployment mechanisms. It provides a set of APIs and configuration options that enable users to easily specify the desired parallelism strategies, communication protocols, and resource allocation policies. Furthermore, PEFT supports various deployment environments, including on-premises clusters, cloud platforms, and specialized hardware accelerators, making it accessible and adaptable to different infrastructure configurations.
\end{itemize}

\subsection{ggml}

The model deployment process typically begins with the testing and creation of the model using a language like Python. However, due to the language's slower execution speed, it is often necessary to convert the model to run on a faster language, such as C++, for deployment. This is where GGML \citep{ggml}, a powerful C++ tensor library, comes into play. GGML not only enables running models natively but also offers additional features like quantization. Notably, GGML is capable of efficiently executing popular models like BLOOM \citep{workshop2023bloom} or Whisper \citep{radford2022robust}, further enhancing its versatility and utility in the deployment process.

\subsection{lamacpp}

Lamacpp \citep{llamacpp} is a highly renowned open-source project focused on enabling the execution of LLaMA (Large Language Model Meta AI) \citep{touvron2023llama} over a terminal interface on a personal computer. This exceptional project encompasses a robust C/C++ implementation that facilitates 4-bit, 5-bit, and 8-bit integer quantization. Notably, Lamacpp has made significant strides by incorporating CUDA and cuBLAS support into its project, further enhancing its capabilities. This project is used as the main test bench to develop new features and improve the GGML library.

\subsection{sherpa}
Sherpa \citep{sherpa} is a powerful implementation of Lamacpp, seamlessly integrated with Flutter, enabling users to run LLaMMa on Windows, Mac, or Android platforms through an intuitive chat interface.

\subsection{mlc-llm}

MCL-LLM \citep{mlc-llm-website} is another solution alternative to sherpa that allows implementing an LLM running locally on multiple platforms. It also offers a framework to allow leveraging the power of local GPUs on a mobile device or laptop for accelerated performance.

\section{LLM training pipeline}
\label{sec:3}

The process of training a Large Language Model involves multiple stages. However, our approach differs from creating a model from scratch. Instead, we opted for a different approach by selecting an existing instruction-based fine-tuned model and improving the fine-tuning process to suit our needs.

\subsection{Choosing a model}

Consideration for choosing a model:
\begin{itemize}
    \item Model Size:
Given that running Large Language Models (LLMs) on Android devices is a relatively new topic, it becomes crucial to consider the hardware specifications of the mobile device. The inference process associated with LLMs is computationally demanding, and the model itself needs to be loaded into RAM. Considering that an Android device consumes around 1.5GB of RAM and the average mobile device has between 4 and 6 GB of RAM we had to consider the model size during the deliberation process to ensure optimal performance. 

\item Quality of the dataset:
The quality of the dataset is of utmost importance when developing any type of deep learning model. The dataset serves as the foundation, and its quality, size, and diversity directly impact the model's capacity to generalize and make accurate predictions.

A high-quality dataset enables the model to understand the problem at hand, facilitating effective learning of patterns, features, and correlations. By exposing the model to diverse examples, the dataset helps it cover different scenarios during the inference process.
Moreover, a good-quality dataset plays a vital role in preventing overfitting or underfitting during the training process. If the dataset is too small or overly specific, it can hinder the model's ability to generalize beyond the limited scope of the data.

    \item  Purpose of the model:
An LLM (Language Model) possesses a wide range of capabilities beyond its text generation capabilities. It can serve diverse purposes, including but not limited to generating code, facilitating engaging chat interactions, and delivering clear instructions.

In our particular endeavor of developing a chat application that triggers action on the mobile device, it was crucial for us to seek out a model specifically designed to excel in chat-based scenarios. We needed a model that could simulate human-like conversations, enabling users to engage in natural and fluid exchanges with the chatbot.
    
\end{itemize}

After careful consideration of the aforementioned aspects, we conducted tests on various models of mobile devices (Android), including Alpaca, Bloomz, Redpajama, and others. Among these options, Redpajama emerged as the most promising due to its relatively small size, with only 3 billion parameters, and its pre-existing fine-tuning that enhances its \textbf{chatting ability}.

We choose Redpajama, developed by RedPajama-INCITE-Chat-3B-v1\citep{huggingface-redpajama-incite-chat} in collaboration with leaders from the open-source AI community, which aims to provide the AI community with open-source models. This particular model contains 2.8 billion parameters and was trained using 131 million tokens.

\subsection{Creating a dataset}

\begin{figure*}
	%\centering
 \includegraphics[width=\linewidth]{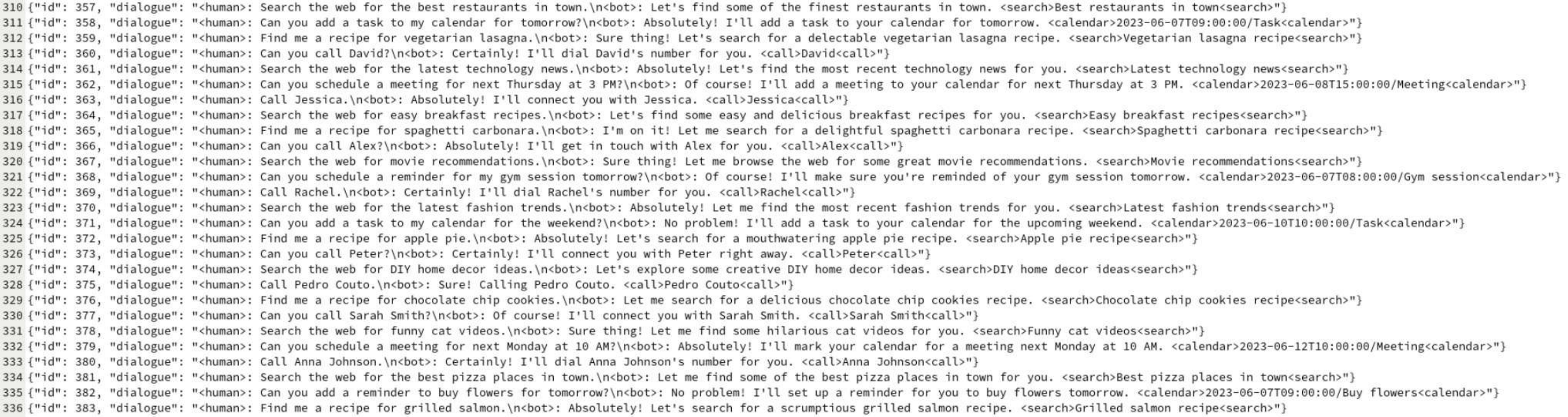}
	\caption{Demo of the dataset generated by GPT3.5}
 \label{fig:demodataset}
\end{figure*}

After conducting a comprehensive analysis of the chosen model, it became evident that a fine-tuning process was essential to elevate the model's performance and align it more closely with our specific needs. Considering that a Language Model (LM) dataset predominantly revolves around textual data, we opted to harness the capabilities of the renowned GPT3.5 model, developed by OpenAI \citep{openai}. Employing the sophisticated technique of prompt engineering, we generated a dataset of a small scale, comprising approximately 357 lines of meticulously crafted data as can be seen in Figure~\ref{fig:demodataset}.

The dataset we constructed primarily revolves around a dialogue between human (<human>) users and the AI bot (<bot>) there were also included special tokens such as <search>, <calendar>, <call> that we will explain later in the paper. The objective is to generate short and succinct responses from the bot, emulating a more conversational and interactive experience. There was also the intention to include normal prompts in the dataset to avoid dataset generalization. By fine-tuning the model using this expansive dataset, we aimed to enhance its contextual understanding, coherence, and overall conversational prowess. To enhance the model's generalization capabilities, we incorporated a blend of action-focused samples and question-answering dialogues encompassing broader knowledge. This approach ensures that the model does not rely solely on specialized tokens, even during regular conversations, thereby preserving its versatility in understanding and responding effectively.

Through the utilization of prompt engineering, we meticulously crafted prompts that encapsulate a wide range of potential user inputs, thereby enabling the bot to generate accurate and relevant responses across diverse conversational scenarios. The careful construction of this dataset ensures that the bot's responses are tailored to the given context, leading to a more engaging and user-friendly interaction.

By incorporating the abundant capabilities of the GPT3.5 model and employing prompt engineering techniques, our fine-tuning process aimed to empower the model to adapt to various prompts, comprehend nuances within the provided text, and generate concise and contextually appropriate responses. This concerted effort ensures that the model is finely tuned to cater to our specific requirements, enabling it to deliver an elevated user experience in terms of dialogue and information exchange.

\subsection{Supervised Fine Tuning}

\begin{figure}
	\centering
\includegraphics[width=\linewidth]{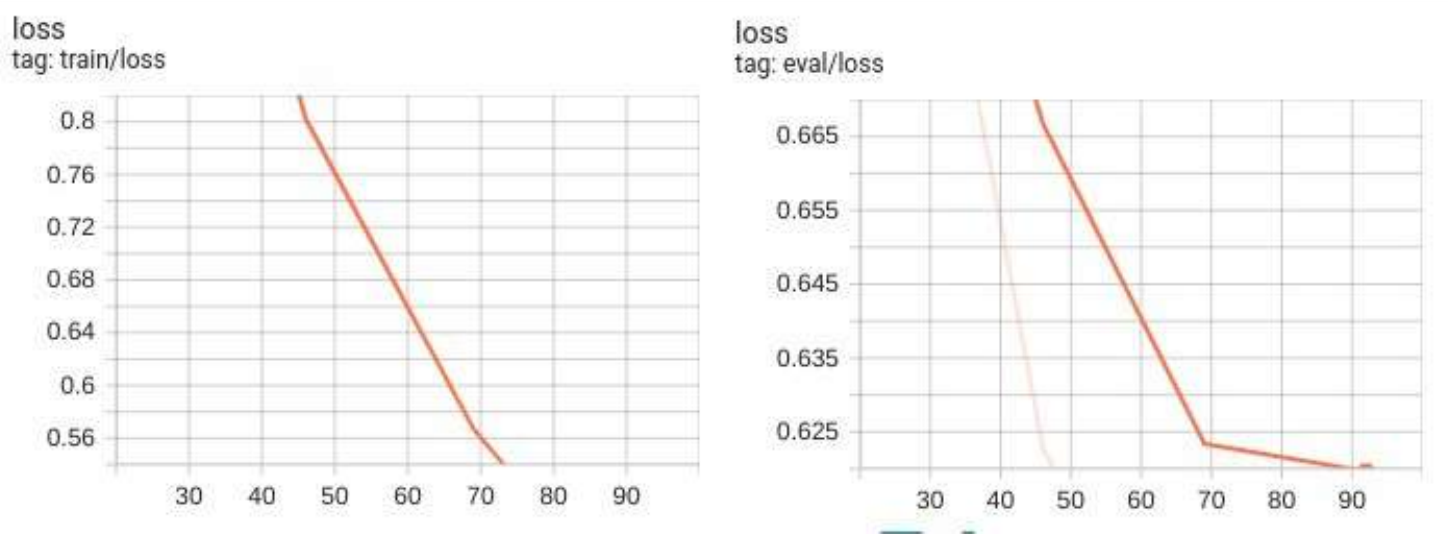}
\caption{Evaluation of the loss in the training and evaluation stages}
\label{fig:trainchart}

\end{figure}

In our approach, we began with the 16-bit base model called RedPajama-INCITE-Chat-3B-v1 (as mentioned previously). To enhance its performance, we utilized a combination of the LoRA and PEFT libraries for fine-tuning. In the LoRA configuration, we specifically selected the parameters (layers) that required fine-tuning while keeping the remainder of the model frozen. Our focus was on fine-tuning the weights associated with the layers responsible for converting embeddings to query keys and values for attention calculations.

We utilized the bitsandbytes library to accelerate the training process and enable 8-bit precision training. We used a conservative approach by adopting a relatively high learning rate of 1e-3 and a small number of epochs (5) due to the small size of our dataset. This decision was based on our observation that, beyond the fifth epoch, the model would start to overfit the training data. We partitioned our dataset into a 90\% training set and a 10\% evaluation set. The progression of loss values for both the training and evaluation sets are visualized in Figure~\ref{fig:trainchart}. To make possible batching samples for faster training we padded the dialogue samples to the length of the biggest dialog with the special end of sentence token (eos).

After completing the training phase, we merged the layers generated by LoRA with the initial base model. Subsequently, we converted this combined 16-bit model into the GGML format, finalizing the enhancement process.

\subsection{Quantization}

Once the training process is completed and we have the trained model in the GGML format, the next crucial step is to further optimize it by quantizing the model. The objective of quantization is to reduce the model's size and hardware requirements without compromising its performance.

To achieve this optimization, we made use of the GGML 4-bit integer quantization feature. By replacing the original 16-bit float weights with 4-bit integers, this strategy effectively reduces the memory footprint and computational complexity of the model's parameters. By quantizing the model, we were able to achieve significant reductions in both the size of the model and the hardware resources needed to deploy and perform inference on it. This is particularly beneficial in scenarios where limited storage or computational power is a constraint.

The quantization process yielded promising results when it comes to model size, which are outlined and discussed in detail in Section~\ref{sec:5}. We couldn't however compare it's performance to the 16-bit model because, at the time of writing this article and to the best of our knowledge, there isn't a way to evaluate a 4-bit quantized model in the GGML format.

\section{Mobile App}
\label{sec:4}

\begin{figure*}
	\centering
\includegraphics[width=0.16\textwidth]{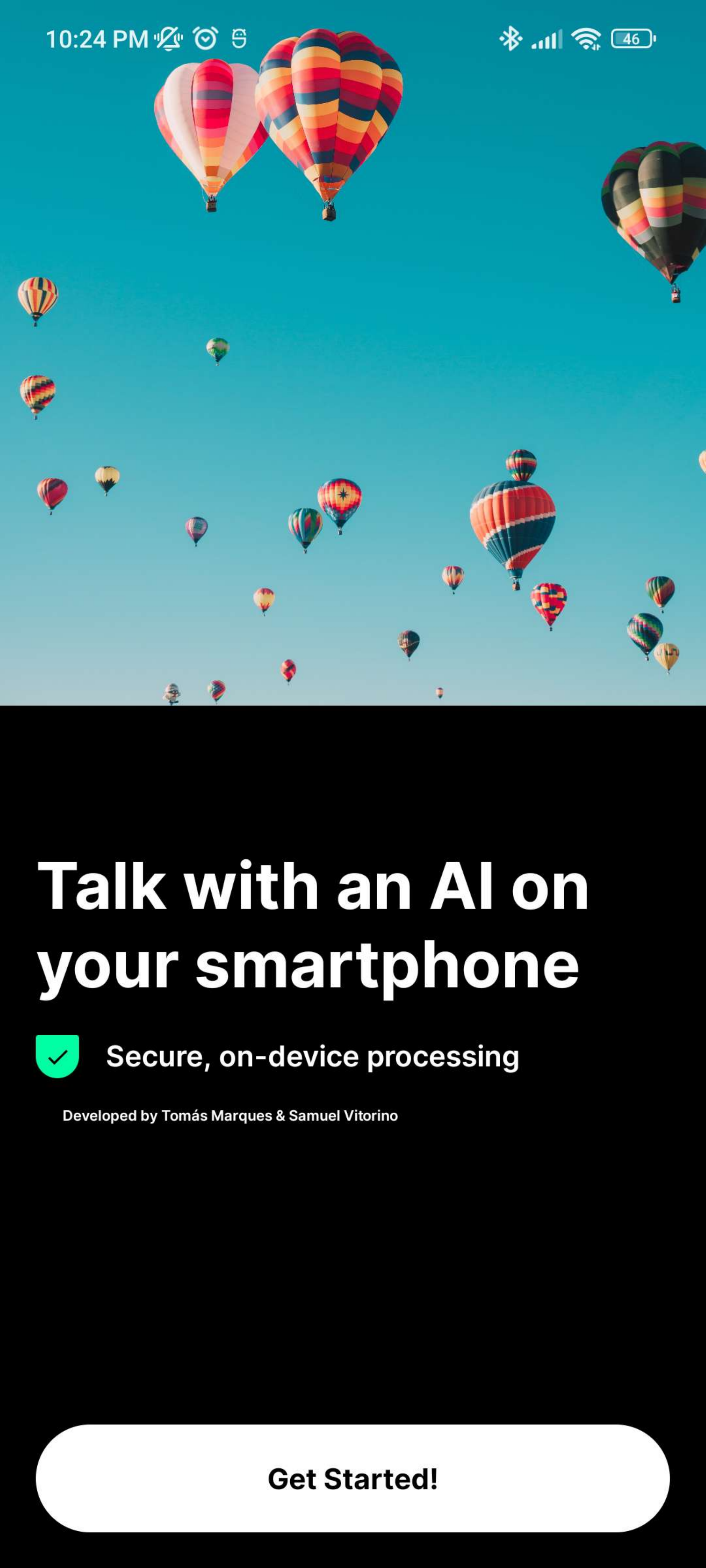}
%\includegraphics[width=0.150\textwidth]{figs/loadinglayers.pdf}
%\label{fig:loadlginlayers}
\includegraphics[width=0.16\textwidth]{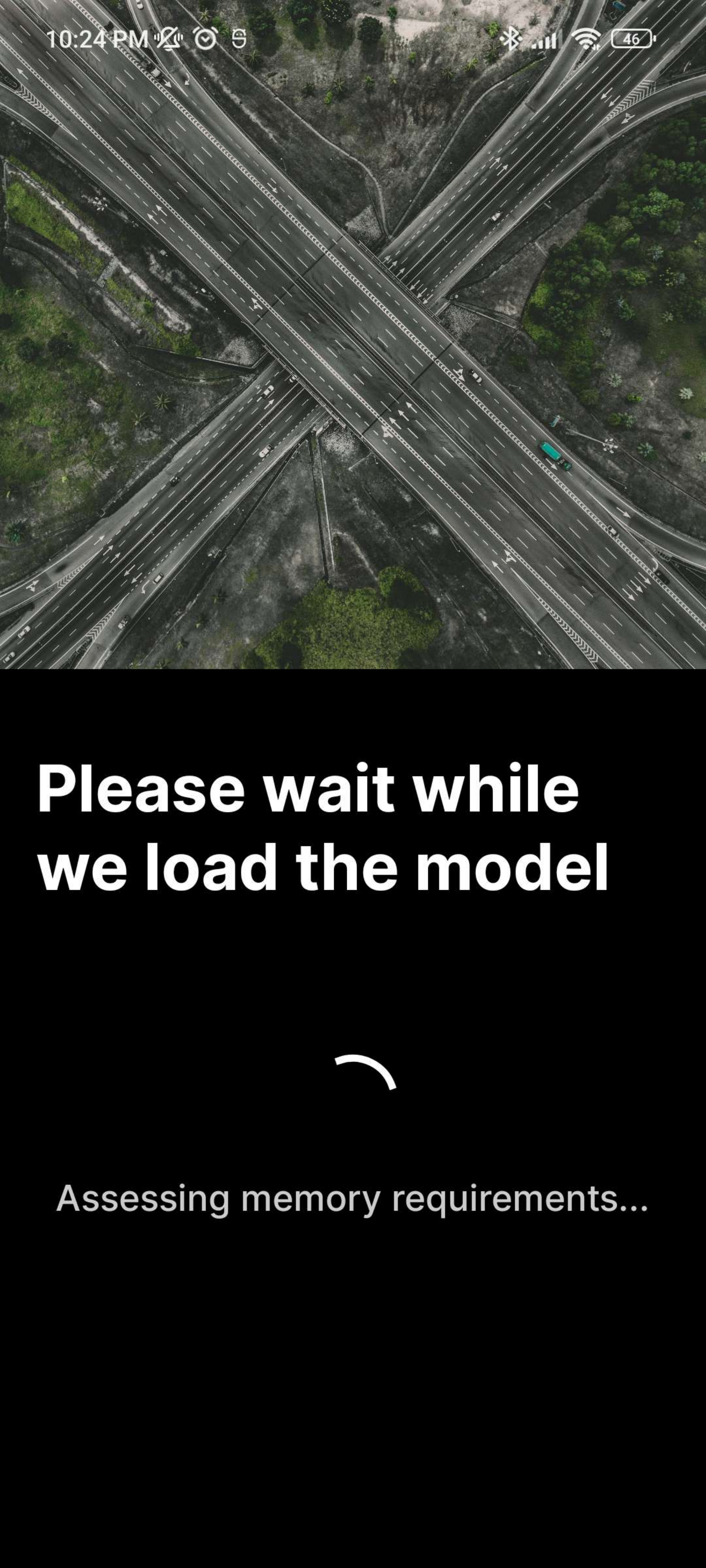}
\includegraphics[width=0.16\textwidth]{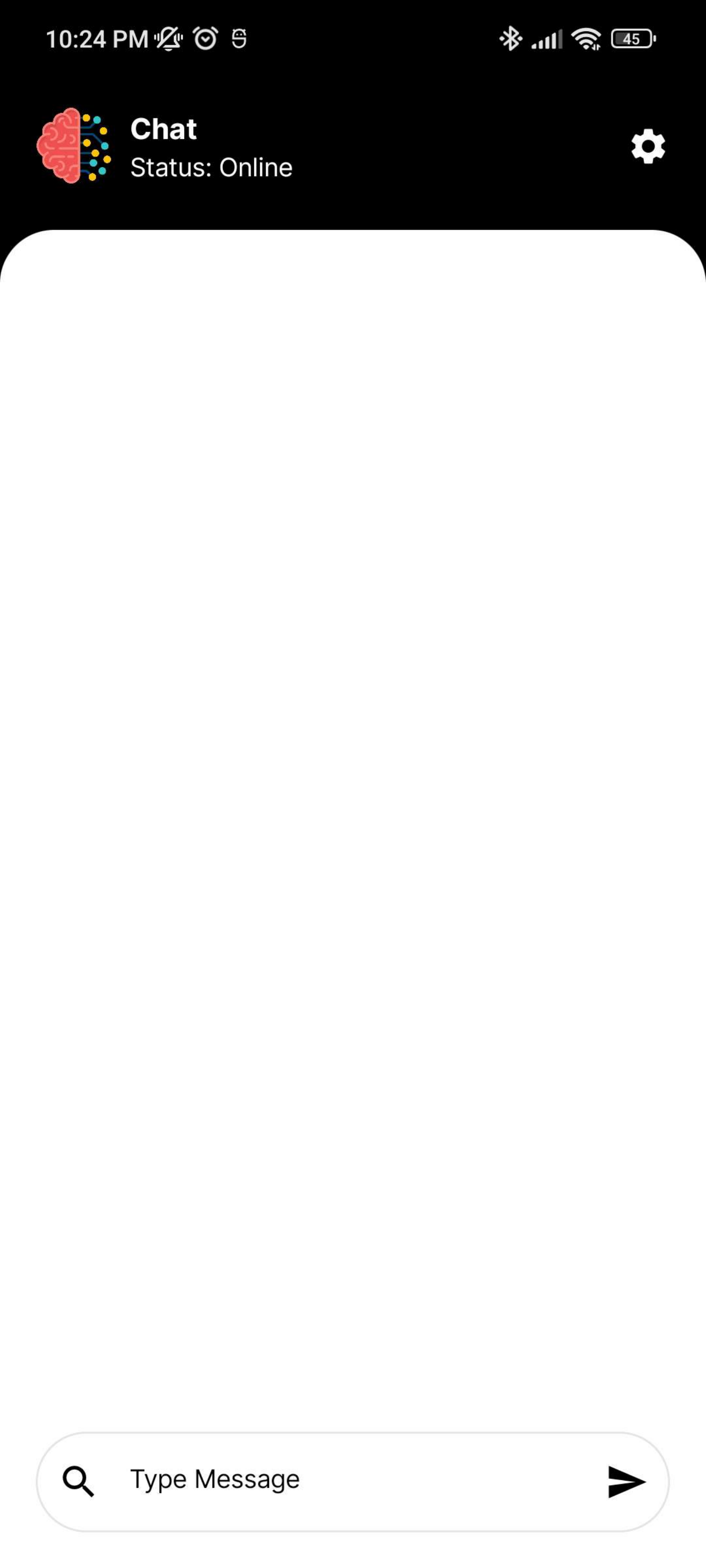}
\includegraphics[width=0.16\textwidth]{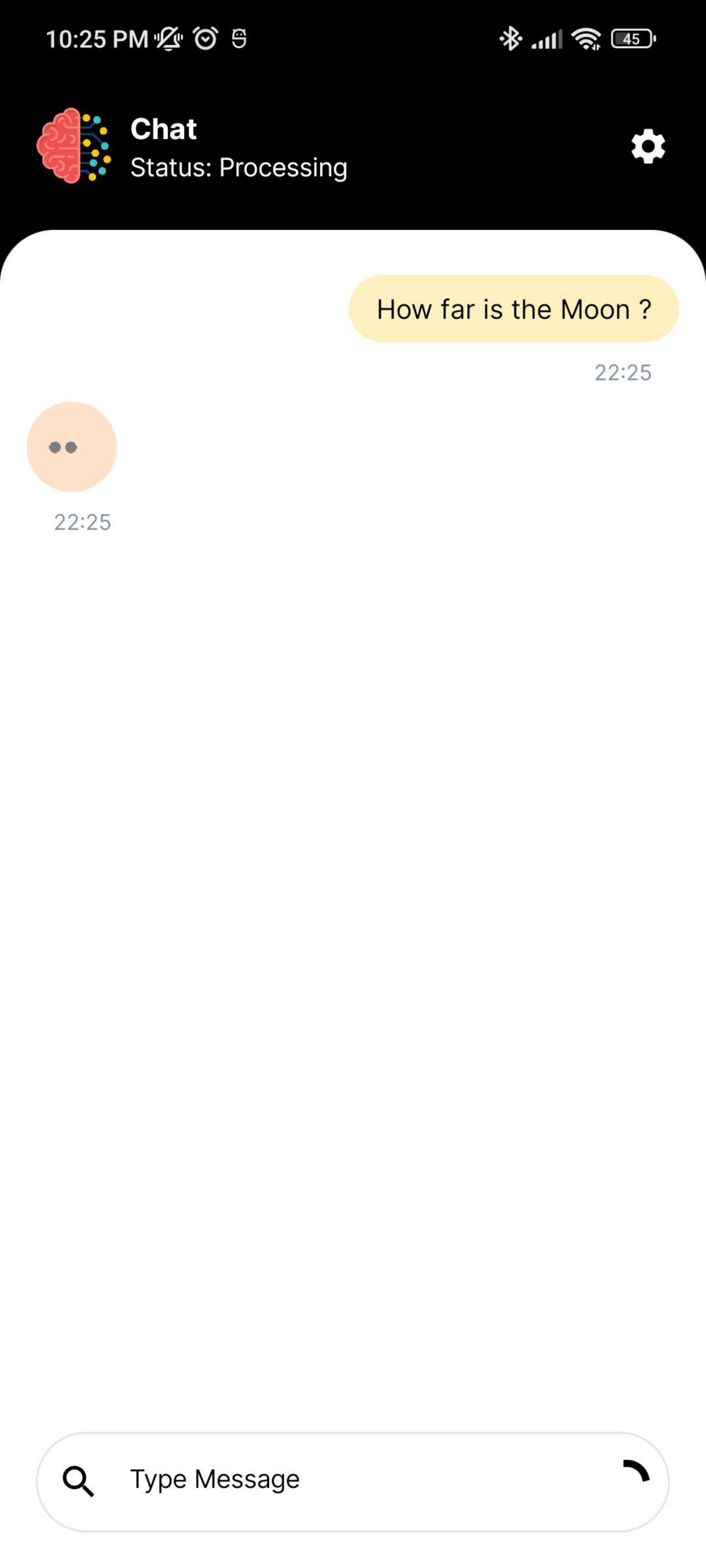}
 \includegraphics[width=0.16\textwidth]{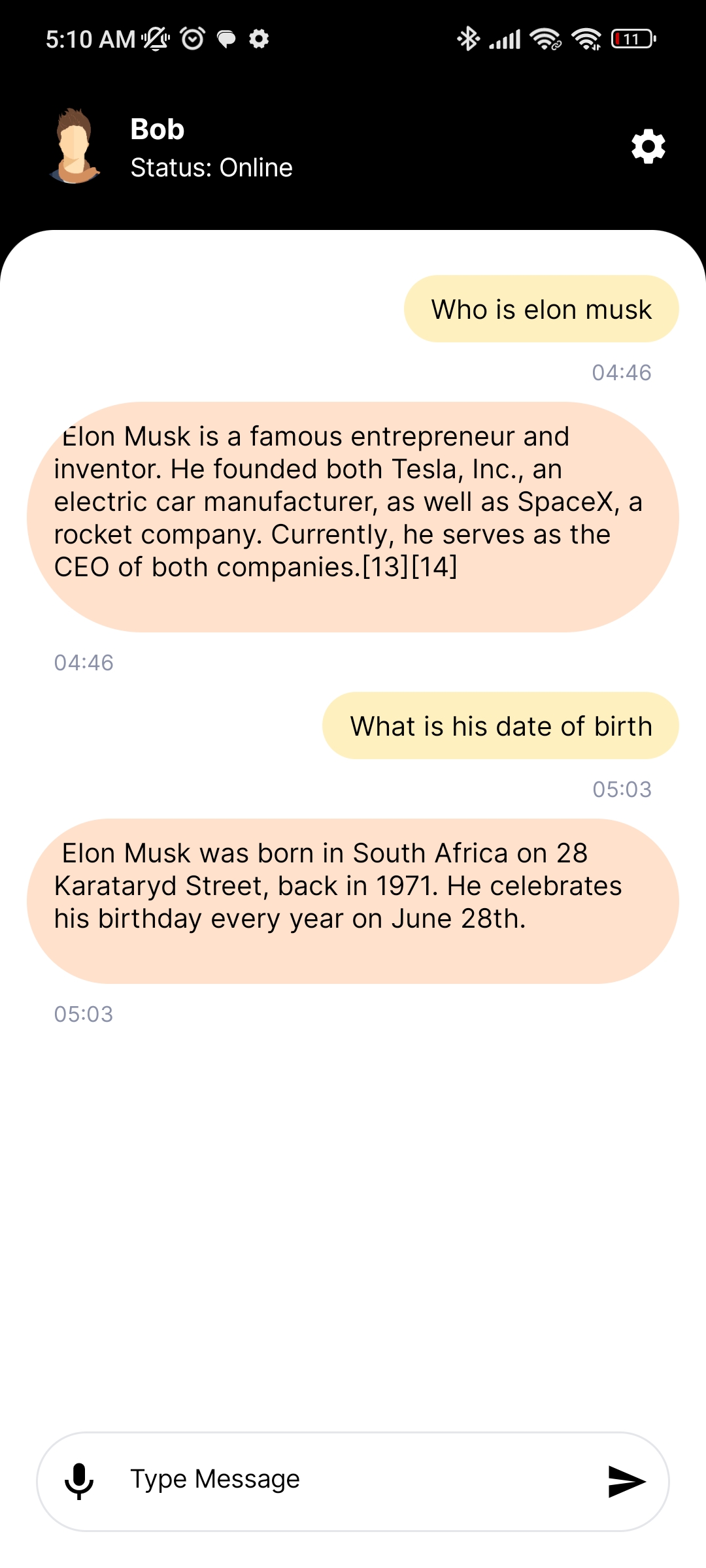}
\includegraphics[width=0.16\textwidth]{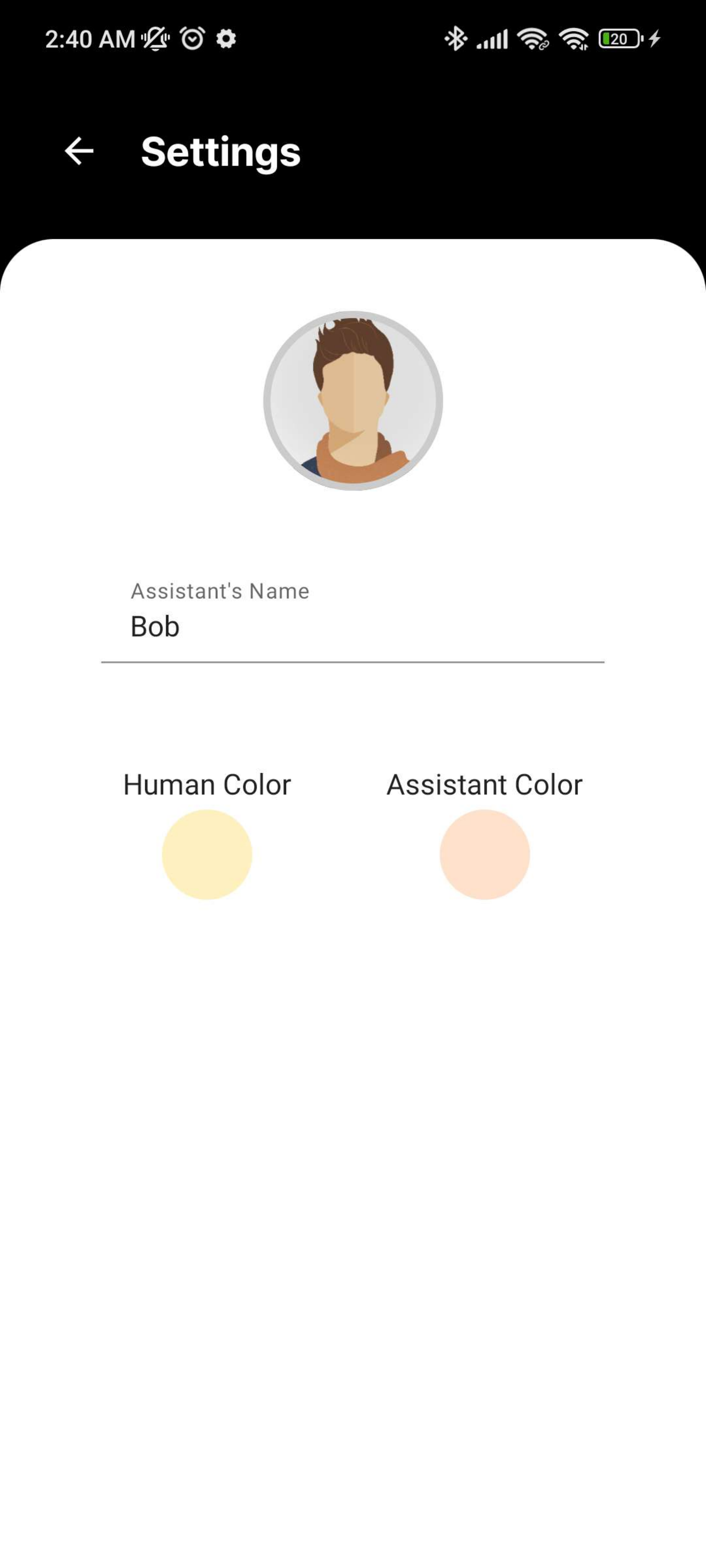}
	\caption{Various Screens of the application: 1-Landing Page, (2)-Loading Page, (3,4,5)-Chat Screen, 6-Settings Screen. In the picture (5) is also possible to visualize the model \textbf{maintaining context} }
\label{fig:screens}
\end{figure*}

 When the user opens the app (Figure~\ref{fig:screens} column 1), they are greeted by a visually captivating landing page designed to instantly capture their attention and create a strong desire to explore the app's features and functionalities but also to them getting the idea of what the app is about just by looking at the page.

After the captivating landing page, the user is seamlessly transitioned to the loading page (Figure~\ref{fig:screens}). Here, the app performs a series of crucial checks to ensure a smooth experience. Firstly, it verifies if the model exists in the phone's data folder. If it does, the app meticulously verifies the file integrity by calculating an MD5 checksum, ensuring the model's integrity.

Upon confirmation of the model's presence, the app proceeds to download it directly from the cloud, ensuring the user has the latest version at their disposal. This streamlined process minimizes any delays or interruptions.

Once the model is successfully downloaded, it is efficiently loaded into the device's memory. This optimization ensures quick access and seamless performance throughout the user's interaction with the app. Once the loading process is complete, the user is gracefully ushered into the chat page, ready to start a conversation.

On the chat page (Figure~\ref{fig:screens} columns 3-5), users are empowered to create prompts in two convenient ways: through typing or by utilizing the microphone button, located at the bottom-left corner of the interface. By simply pressing the mic button, users can effortlessly convert their speech into text, enhancing accessibility and convenience even if the device is offline.

When a prompt is sent, it triggers the inference process within the app. As the app diligently generates tokens and presents them, users can witness the app's intelligent responses unfold in real time. After the prompt is sent the app enters the loading phase where the user can not send any more prompts until the model is generating tokens (Figure~\ref{fig:screens} column 4). 

This allows the LLM to maintain a context of the conversation. Maintaining context as can be seen in Figure~\ref{fig:screens} (column 5) where the model is capable of remembering that the conversation is about Elon Musk without  the need to repeat it in the second prompt.

At any time the user can enter the settings screen (Figure~ \ref{fig:screens} column 6) where he can customize the profile picture, username and the color of the chat individually. 
 
\subsection{Embedding an LLM on a Mobile App}

\begin{figure}
	%\centering
 \includegraphics[width=\linewidth]{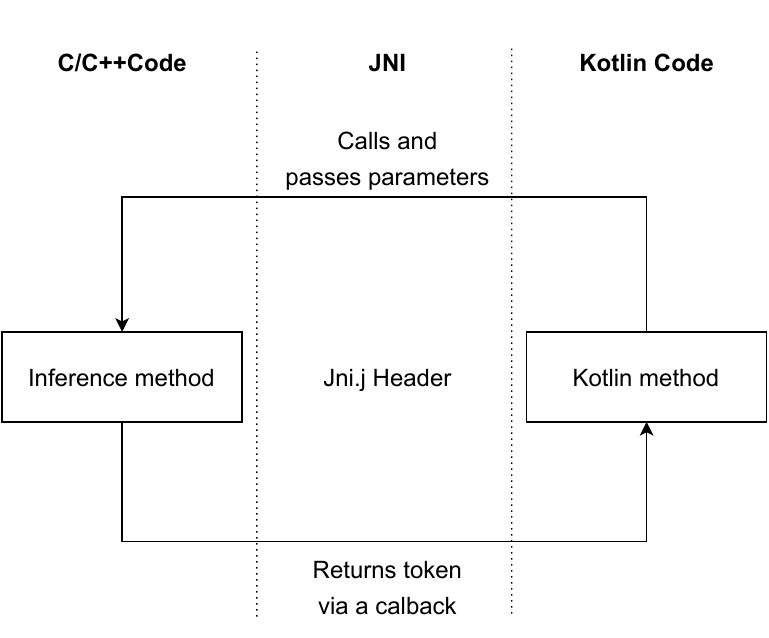}
	\caption{Android NDK architecture}
 \label{fig:ndkarc}
\end{figure}

Our approach to run the inference model on an Android device was to use transformer architecture for GPT-Neox provided by GGML \citep{gptneox}. This implementation is developed in C++ and proved to be highly effective in our use case. To allow the process of compiling the code and run inference the inference method on the model we use used Android NDK(Native Development Kit) to compile and run in Android.

Android NDK is a set of tools provided by directly by googled its main objective was to allow the developer to write performance-critical portions of code in C and C++. 

To provide a visual representation, please refer to the diagram in Figure~\ref{fig:ndkarc} illustrating the architecture of the Android NDK in action. As depicted, the NDK operates within a separate thread, allowing it to execute inference calls with their respective parameters. 

Upon completion, the inference process returns the generated token through a callback method. This callback mechanism allows for seamless integration with the rest of the application, enabling further processing or presentation of the obtained token.

This integration of our model, the Android NDK, and our Android device serves as a powerful combination, facilitating the execution of complex inference tasks seamlessly on mobile devices.

\subsection{From text to actions}
\label{sub:fromtext}

To fully leverage the capabilities of mobile devices and enhance app functionality, we devised a method for our Language Learning Model (LLM) to interact with phone features. This involved incorporating special tags into our dataset to perform specific actions. In this application version, we have focused on supporting three key actions: making calls, scheduling events in the calendar, and performing web searches.

\begin{figure}
\centering
 \includegraphics[width=0.15\textwidth]{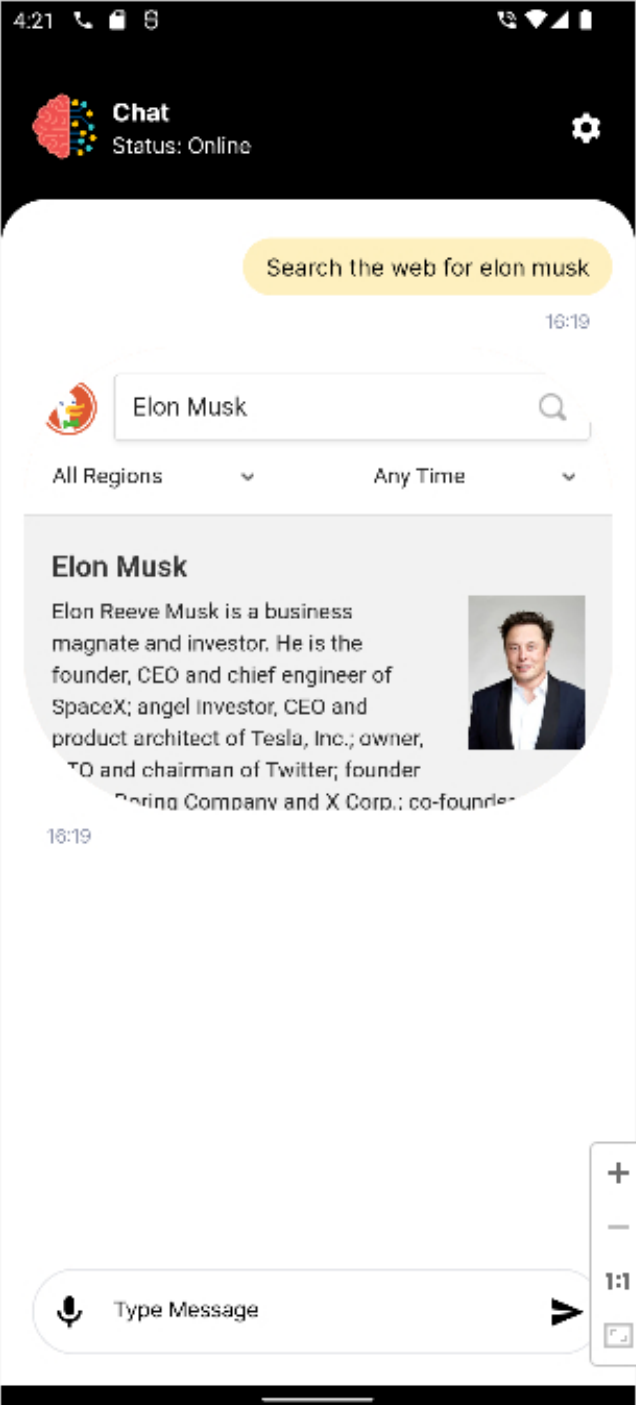}
 \includegraphics[width=0.15\textwidth]{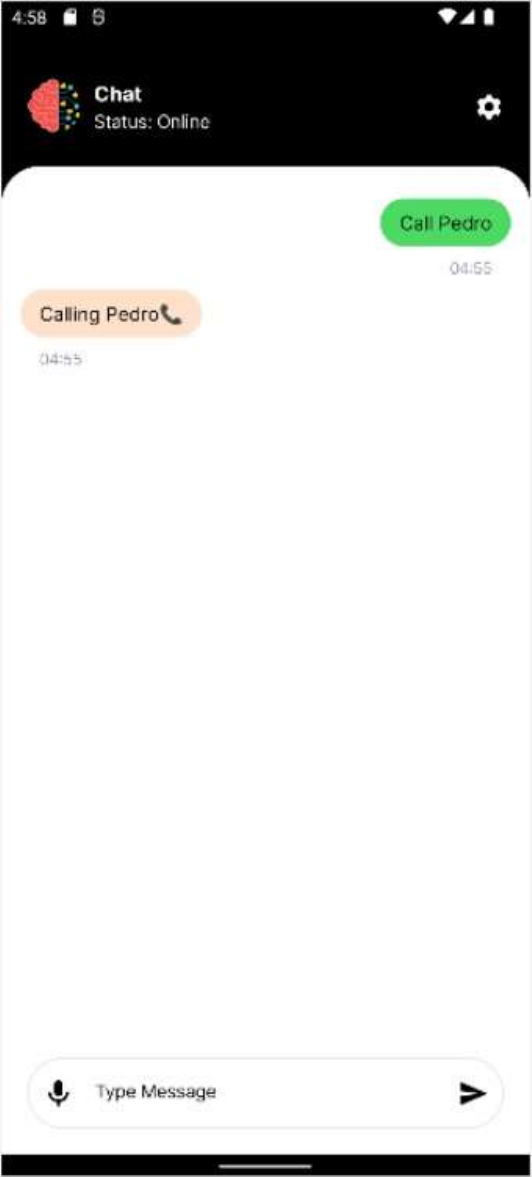}
 \includegraphics[width=0.15\textwidth]{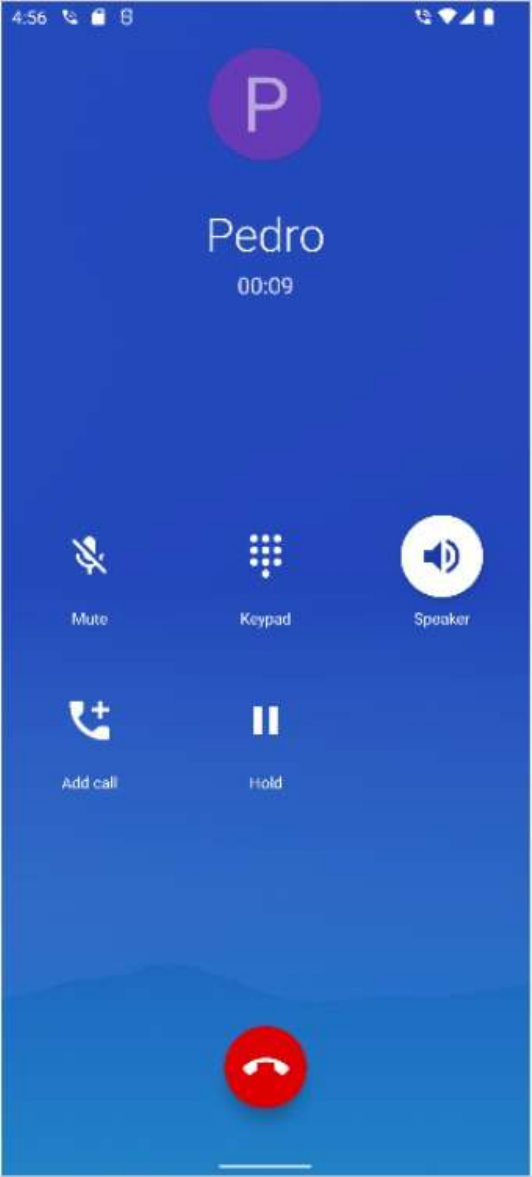}
	\caption{Example of the search the web and call action features}
 \label{fig:elon}
\end{figure}

To initiate a call (Figure~\ref{fig:elon} second and third column), the model simply needs to generate the special token "<call>" followed by the contact's name, and then conclude with another special token "<call>". For instance, "<call>John<call>" would instruct the model to initiate a call to John.

Similarly, the process for conducting a web search (Figure~\ref{fig:elon} first column) is analogous. The model needs to generate the special token "<search>", followed by the desired query, and then conclude with the same special token "<search>". For example, "<search>Highest building in the world<search>" would prompt the model to perform a web search for the tallest building on earth.

Finally, the model must generate the special token "<calendar>", then the event's date, time, and name, all of which are separated by a slash ("/"), before concluding with the special token "calendar>". For example, "<calendar>2023-05-20T09:00:00/Meeting<calendar>" would instruct the model to plan a meeting on May 20, 2023, at 9:00 AM.

With the use of this powerful feature, which we call \textbf{text-to-actions}, the model's capabilities may be extended beyond simple text generation and use a variety of mobile device features.

\subsection{Limitations}

Running a Large Language Model (LLM) on a mobile device presents several limitations:
\begin{itemize}
    \item Computational Resources: LLMs require substantial computational resources, including processing power and memory. Mobile devices often have limited resources compared to desktop computers or servers. As a result, running LLMs on mobile devices may lead to slower inference times and reduced overall performance.
    \item Memory Constraints: LLMs tend to have large model sizes, often ranging from hundreds of megabytes to several gigabytes. Mobile devices typically have limited storage and RAM capacities. Loading and storing such large models can strain the device's memory, potentially leading to slower operation and higher memory usage.
    \item Power Consumption: LLMs are computationally intensive and can consume a significant amount of power during inference. Mobile devices have limited battery capacity, and running LLMs for extended periods can drain the battery quickly. This limitation restricts the practicality and usability of LLMs on mobile devices, particularly when offline or in situations without immediate access to power sources.
\end{itemize}

\section{Tests and Results}
\label{sec:5}
We ran some tests on physical devices and got to the conclusion that the 6GB Android device offers a reasonable level of performance, while the 4GB device represents the bare minimum required to ensure the application functions properly. 
We reached the conclusion through some tests that the app maintains conversation context.
When it comes to the user interface (UI), we were pleasantly surprised by the design interface. It surpassed our expectations by being exceptionally intuitive, making it incredibly easy for users to navigate and interact with the app. The thoughtfully designed UI elements, including clear visual cues and logical placement of controls, contribute to an enjoyable user experience.
One of the standout features of the app is its voice functionality. We were particularly impressed with its performance, as it consistently delivered excellent results. Notably, the voice method even works offline, providing users with uninterrupted access to voice-based interactions regardless of their internet connectivity. This offline capability sets the app apart and enhances its versatility in various scenarios.
\begin{table}[width=.9\linewidth,cols=3,pos=h]
\caption{List of android devices tested.}\label{table:hardwaretests}
\begin{tabular*}{\tblwidth}{@{} LLL@{} }
\toprule
Android Model & CPU & RAM\\
\midrule
POCO X3 PRO & 4core 2.15ghz & 6GB \\
LeMax2 & 8core 2.96ghz & 4GB \\

%FALAR SOBRE END OF SENTENCE

\bottomrule
\end{tabular*}
\end{table}

\subsection{Quantization results}

\begin{table}[width=.9\linewidth,cols=3,pos=h]
\caption{Size of the initial 16-bit model and the 4-bit quantized model}\label{table:hardwaretests}
\label{tab:quant}
\begin{tabular*}{\tblwidth}{@{} LLL@{} }
\toprule
Model & Precision & Size\\
\midrule
st-ggml-model-16bit & float16 & 5.17GB \\
st-ggml-model-q4\_0 & int4 & \textbf{1.6GB} \\

%FALAR SOBRE END OF SENTENCE

\bottomrule
\end{tabular*}
\end{table}

Following the successful quantization process, our final model has undergone a remarkable transformation, boasting an impressive size reduction of 70\% (Table~\ref{tab:quant}). This significant achievement empowers us to deploy the model effortlessly on mobile devices, even those equipped with a modest 4GB of memory. By harnessing the power of quantization, we have not only optimized the model's performance but also paved the way for seamless utilization in resource-constrained environments. 

\subsection{Text to actions tests}

The base model exhibits certain issues when determining where to appropriately conclude the text and generate the designated special token "<|endoftext|>". We are uncertain whether this behavior arises due to its training methodology, but as a result, the model frequently reproduces the human prompt and struggles to halt at the intended position. Additionally, we have encountered difficulties with the quantized model, whereby it occasionally struggles to identify the correct token (tag) for the desired action. Even when it does select the appropriate option, it sometimes mistakenly chooses an incorrect closing special token, further complicating the output. For example, the model can generate an output such as "<call>John Castro<calendar>".

\section{Conclusion}
\label{sec:6}

In conclusion, the field of Artificial Intelligence has witnessed significant advancements in large language models, particularly with the emergence of transformer architecture and the development of models like ChatGPT by OpenAI. While leveraging powerful cloud servers for processing tasks offers advantages such as faster inference capabilities, it also presents challenges in terms of latency and privacy. To address these concerns, running LLMs directly on edge devices represents the future of this technology.

This article has presented a groundbreaking approach to on-device inference of GPT LLMs, showcasing the potential of real-time execution without the need for a network connection. The integration of native code and model quantization techniques enables the smooth operation of fine-tuned GPT LLMs with billions of parameters on devices with limited memory. The application developed in this study not only serves as a general-purpose assistant but also facilitates seamless mobile phone interactions for various tasks.

Future directions for this technology involve further advancements and optimizations to make powerful LLMs accessible and seamlessly integrated into everyday devices. With the increasing prevalence of custom accelerators in mobile devices, the vision of executing massive LLMs on the device without network dependency is within reach. This advancement has the potential to revolutionize the way we interact with AI systems, empowering users with sophisticated AI capabilities at their fingertips.

On-device inference of LLMs holds immense promise, and its realization will lead to transformative advancements in AI technology. By continuing to push the boundaries and refine the methodologies, we can create a future where powerful LLMs are seamlessly integrated into everyday devices, enhancing user experiences and providing ubiquitous access to AI capabilities.

\bibliographystyle{cas-model2-names}

\cite{*}
\bibliography{refs}

\begin{thebibliography}{15}
\expandafter\ifx\csname natexlab\endcsname\relax\def\natexlab#1{#1}\fi
\providecommand{\url}[1]{\texttt{#1}}
\providecommand{\href}[2]{#2}
\providecommand{\path}[1]{#1}
\providecommand{\DOIprefix}{doi:}
\providecommand{\ArXivprefix}{arXiv:}
\providecommand{\URLprefix}{URL: }
\providecommand{\Pubmedprefix}{pmid:}
\providecommand{\doi}[1]{\href{http://dx.doi.org/#1}{\path{#1}}}
\providecommand{\Pubmed}[1]{\href{pmid:#1}{\path{#1}}}
\providecommand{\bibinfo}[2]{#2}
\ifx\xfnm\relax \def\xfnm[#1]{\unskip,\space#1}\fi
%Type = Misc
\bibitem[{Bip-Rep(2023)}]{sherpa}
\bibinfo{author}{Bip-Rep}, \bibinfo{year}{2023}.
\newblock \bibinfo{title}{{sherpa}}.
\newblock \bibinfo{howpublished}{\url{https://github.com/Bip-Rep/sherpa}}.
\newblock \bibinfo{note}{Accessed: Date}.
%Type = Misc
\bibitem[{{Gerganov, Georgi}(2023)}]{ggml}
\bibinfo{author}{{Gerganov, Georgi}}, \bibinfo{year}{2023}.
\newblock \bibinfo{title}{Ggml}.
\newblock \bibinfo{howpublished}{\url{https://github.com/ggerganov/ggml}}.
%Type = Misc
\bibitem[{ggerganov(2023a)}]{gptneox}
\bibinfo{author}{ggerganov}, \bibinfo{year}{2023}a.
\newblock \bibinfo{title}{{ggerganov/ggml/gpt-neox}}.
\newblock
  \bibinfo{howpublished}{\url{https://github.com/ggerganov/ggml/tree/master/examples/gpt-neox}}.
\newblock \bibinfo{note}{Accessed: June 14, 2023}.
%Type = Misc
\bibitem[{ggerganov(2023b)}]{llamacpp}
\bibinfo{author}{ggerganov}, \bibinfo{year}{2023}b.
\newblock \bibinfo{title}{{llama.cpp}}.
\newblock \bibinfo{howpublished}{\url{https://github.com/ggerganov/llama.cpp}}.
%Type = Misc
\bibitem[{Hu et~al.(2021)Hu, Shen, Wallis, Allen-Zhu, Li, Wang, Wang and
  Chen}]{hu2021lora}
\bibinfo{author}{Hu, E.J.}, \bibinfo{author}{Shen, Y.},
  \bibinfo{author}{Wallis, P.}, \bibinfo{author}{Allen-Zhu, Z.},
  \bibinfo{author}{Li, Y.}, \bibinfo{author}{Wang, S.}, \bibinfo{author}{Wang,
  L.}, \bibinfo{author}{Chen, W.}, \bibinfo{year}{2021}.
\newblock \bibinfo{title}{Lora: Low-rank adaptation of large language models}.
\newblock \href{http://arxiv.org/abs/2106.09685}{\tt arXiv:2106.09685}.
%Type = Misc
\bibitem[{{Hugging Face}(2023)}]{huggingface-peft}
\bibinfo{author}{{Hugging Face}}, \bibinfo{year}{2023}.
\newblock \bibinfo{title}{{PEFT}: Practical entity framework for transformer}.
\newblock \bibinfo{howpublished}{\url{https://github.com/huggingface/peft}}.
%Type = Article
\bibitem[{Lin et~al.(2023)Lin, Tang, Tang, Yang, Dang and Han}]{lin2023awq}
\bibinfo{author}{Lin, J.}, \bibinfo{author}{Tang, J.}, \bibinfo{author}{Tang,
  H.}, \bibinfo{author}{Yang, S.}, \bibinfo{author}{Dang, X.},
  \bibinfo{author}{Han, S.}, \bibinfo{year}{2023}.
\newblock \bibinfo{title}{Awq: Activation-aware weight quantization for llm
  compression and acceleration}.
\newblock \bibinfo{journal}{arXiv preprint arXiv:2306.00978} .
%Type = Misc
\bibitem[{{MLC.AI}(2023)}]{mlc-llm-website}
\bibinfo{author}{{MLC.AI}}, \bibinfo{year}{2023}.
\newblock \bibinfo{title}{{mlc-llm}}.
\newblock \bibinfo{howpublished}{\url{https://mlc.ai/mlc-llm/}}.
\newblock \bibinfo{note}{Accessed: Date}.
%Type = Article
\bibitem[{OpenAI(2021)}]{openai}
\bibinfo{author}{OpenAI}, \bibinfo{year}{2021}.
\newblock \bibinfo{title}{Openai: An ai research laboratory} \URLprefix
  \url{https://openai.com/}.
%Type = Misc
\bibitem[{OpenAI(2023)}]{openai2023gpt4}
\bibinfo{author}{OpenAI}, \bibinfo{year}{2023}.
\newblock \bibinfo{title}{Gpt-4 technical report}.
\newblock \href{http://arxiv.org/abs/2303.08774}{\tt arXiv:2303.08774}.
%Type = Misc
\bibitem[{Radford et~al.(2022)Radford, Kim, Xu, Brockman, McLeavey and
  Sutskever}]{radford2022robust}
\bibinfo{author}{Radford, A.}, \bibinfo{author}{Kim, J.W.},
  \bibinfo{author}{Xu, T.}, \bibinfo{author}{Brockman, G.},
  \bibinfo{author}{McLeavey, C.}, \bibinfo{author}{Sutskever, I.},
  \bibinfo{year}{2022}.
\newblock \bibinfo{title}{Robust speech recognition via large-scale weak
  supervision}.
\newblock \href{http://arxiv.org/abs/2212.04356}{\tt arXiv:2212.04356}.
%Type = Misc
\bibitem[{togethercomputer(2023)}]{huggingface-redpajama-incite-chat}
\bibinfo{author}{togethercomputer}, \bibinfo{year}{2023}.
\newblock \bibinfo{title}{{RedPajama-INCITE-Chat-3B-v1} model on hugging face}.
\newblock \bibinfo{howpublished}{Hugging Face}.
\newblock \URLprefix
  \url{https://huggingface.co/togethercomputer/RedPajama-INCITE-Chat-3B-v1}.
%Type = Misc
\bibitem[{Touvron et~al.(2023)Touvron, Lavril, Izacard, Martinet, Lachaux,
  Lacroix, Rozière, Goyal, Hambro, Azhar, Rodriguez, Joulin, Grave and
  Lample}]{touvron2023llama}
\bibinfo{author}{Touvron, H.}, \bibinfo{author}{Lavril, T.},
  \bibinfo{author}{Izacard, G.}, \bibinfo{author}{Martinet, X.},
  \bibinfo{author}{Lachaux, M.A.}, \bibinfo{author}{Lacroix, T.},
  \bibinfo{author}{Rozière, B.}, \bibinfo{author}{Goyal, N.},
  \bibinfo{author}{Hambro, E.}, \bibinfo{author}{Azhar, F.},
  \bibinfo{author}{Rodriguez, A.}, \bibinfo{author}{Joulin, A.},
  \bibinfo{author}{Grave, E.}, \bibinfo{author}{Lample, G.},
  \bibinfo{year}{2023}.
\newblock \bibinfo{title}{Llama: Open and efficient foundation language
  models}.
\newblock \href{http://arxiv.org/abs/2302.13971}{\tt arXiv:2302.13971}.
%Type = Misc
\bibitem[{Vaswani et~al.(2017)Vaswani, Shazeer, Parmar, Uszkoreit, Jones,
  Gomez, Kaiser and Polosukhin}]{vaswani2017attention}
\bibinfo{author}{Vaswani, A.}, \bibinfo{author}{Shazeer, N.},
  \bibinfo{author}{Parmar, N.}, \bibinfo{author}{Uszkoreit, J.},
  \bibinfo{author}{Jones, L.}, \bibinfo{author}{Gomez, A.N.},
  \bibinfo{author}{Kaiser, L.}, \bibinfo{author}{Polosukhin, I.},
  \bibinfo{year}{2017}.
\newblock \bibinfo{title}{Attention is all you need}.
\newblock \href{http://arxiv.org/abs/1706.03762}{\tt arXiv:1706.03762}.
%Type = Misc
\bibitem[{Workshop(2023)}]{workshop2023bloom}
\bibinfo{author}{Workshop, B.}, \bibinfo{year}{2023}.
\newblock \bibinfo{title}{Bloom: A 176b-parameter open-access multilingual
  language model}.
\newblock \href{http://arxiv.org/abs/2211.05100}{\tt arXiv:2211.05100}.

\end{thebibliography}


\bio{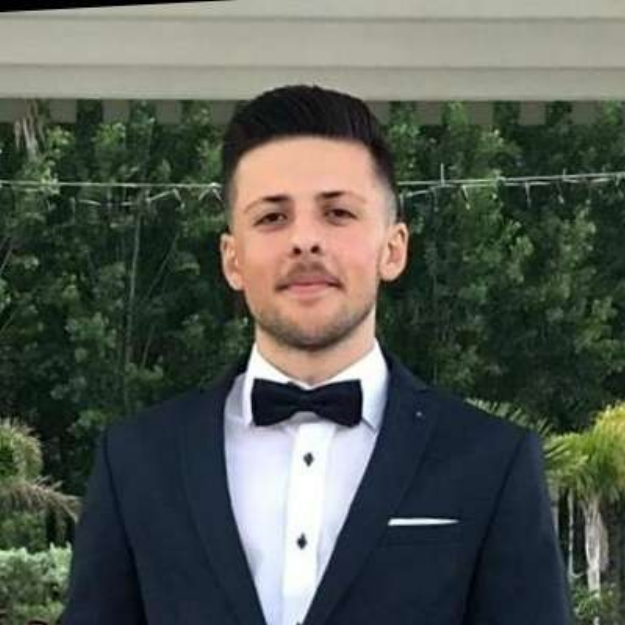}
Hi, my name is Tomás Marques, im 22 years old, I like to code on Linux, and listen to music. Currently, I'm MSc student and Researcher in Computer Engineering @Polytechnic of Leiria.\\\\\\\\

\endbio

\bio{figs/pic1}
My name is Samuel Carreira, and I am 21 years old. Currently, I'm doing a Master's in Computer Engineering - Mobile Computing at the Polytechnic of Leiria. I'm also a researcher in Artificial Intelligence and Computer Vision.

\endbio

\end{document}